\documentclass{article}
\usepackage[margin=2.5cm]{geometry}
\usepackage{graphicx}
\usepackage{picinpar}
\usepackage{subcaption}
\usepackage{colortbl}
\usepackage{booktabs}
\usepackage{multirow}
\usepackage{soul}
\usepackage{enumerate}
\usepackage{color}
\usepackage{hyperref}
\usepackage{amsmath}
\usepackage{amssymb}
\usepackage{algorithm}
\usepackage{algpseudocode}
\usepackage{kpfonts}

\usepackage{times}
\usepackage{latexsym}
\usepackage{xcolor}

\usepackage{MnSymbol}

\usepackage{fancyhdr}
\providecommand{\keywords}[1]
{
  \small	
  \textbf{\textit{Keywords---}} #1
}

\begin{document}
\title{Multitask Recalibrated Aggregation Network \\ for Medical Code Prediction}
%
%
\author{
\ Wei Sun$^{\dag}$, Shaoxiong Ji$^{\dag}\thanks{Corresponding author: Shaoxiong Ji.}~$, Erik Cambria$^{\ddag}$ and Pekka Marttinen$^{\dag}$\\
$^{\dag}$ Aalto University \\
\texttt{\{wei.sun; shaoxiong.ji; pekka.marttinen\}@aalto.fi} \\
$^{\ddag}$ Nanyang Technological University \\
\texttt{cambria@ntu.edu.sg} \\
}
\date{}
%
%
%
\maketitle              

\thispagestyle{fancy}

\begin{abstract}
Medical coding translates professionally written medical reports into standardized codes, which is an essential part of medical information systems and health insurance reimbursement. 
Manual coding by trained human coders is time-consuming and error-prone. 
Thus, automated coding algorithms have been developed, building especially on the recent advances in machine learning and deep neural networks.
To solve the challenges of encoding lengthy and noisy clinical documents and capturing code associations, we propose a multitask recalibrated aggregation network.
In particular, multitask learning shares information across different coding schemes and captures the dependencies between different medical codes. Feature recalibration and aggregation in shared modules enhance representation learning for lengthy notes.
Experiments with a real-world MIMIC-III dataset show significantly improved predictive performance. 

\end{abstract}

\section{Introduction}

Clinical notes generated by clinicians contain rich information about patients' diagnoses and treatment procedures. 
Healthcare institutions digitized these clinical texts into Electronic Health Records (EHRs), together with other structural medical and treatment histories of patients, for clinical data management, health condition tracking and automation.
To facilitate information management, clinical notes are usually annotated with standardized statistical codes. 
Different diagnosis classification systems utilize various medical coding systems.  
One of the most widely used coding systems is the International Classification of Diseases (ICD) maintained by the World Health Organization\footnote{\url{https://www.who.int/standards/classifications/classification-of-diseases}}.
The ICD system is used to transform diseases, symptoms, signs, and treatment procedures into standard medical codes and has been widely used for clinical data analysis, automated medical decision support~\cite{choi2016doctor}, and medical insurance reimbursement~\cite{park2000accuracy}. 
The latest ICD version is ICD-11 that will become effective in 2022, while older versions such as ICD-9 and ICD-9-CM, ICD-10 are also concurrently used. 
Other popular medical condition classification tools include the Clinical Classifications Software (CCS) and Hierarchical Condition Category (HCC) coding. 

\begin{figure*}[htbp]
\centering
	\includegraphics[width=0.8\textwidth]{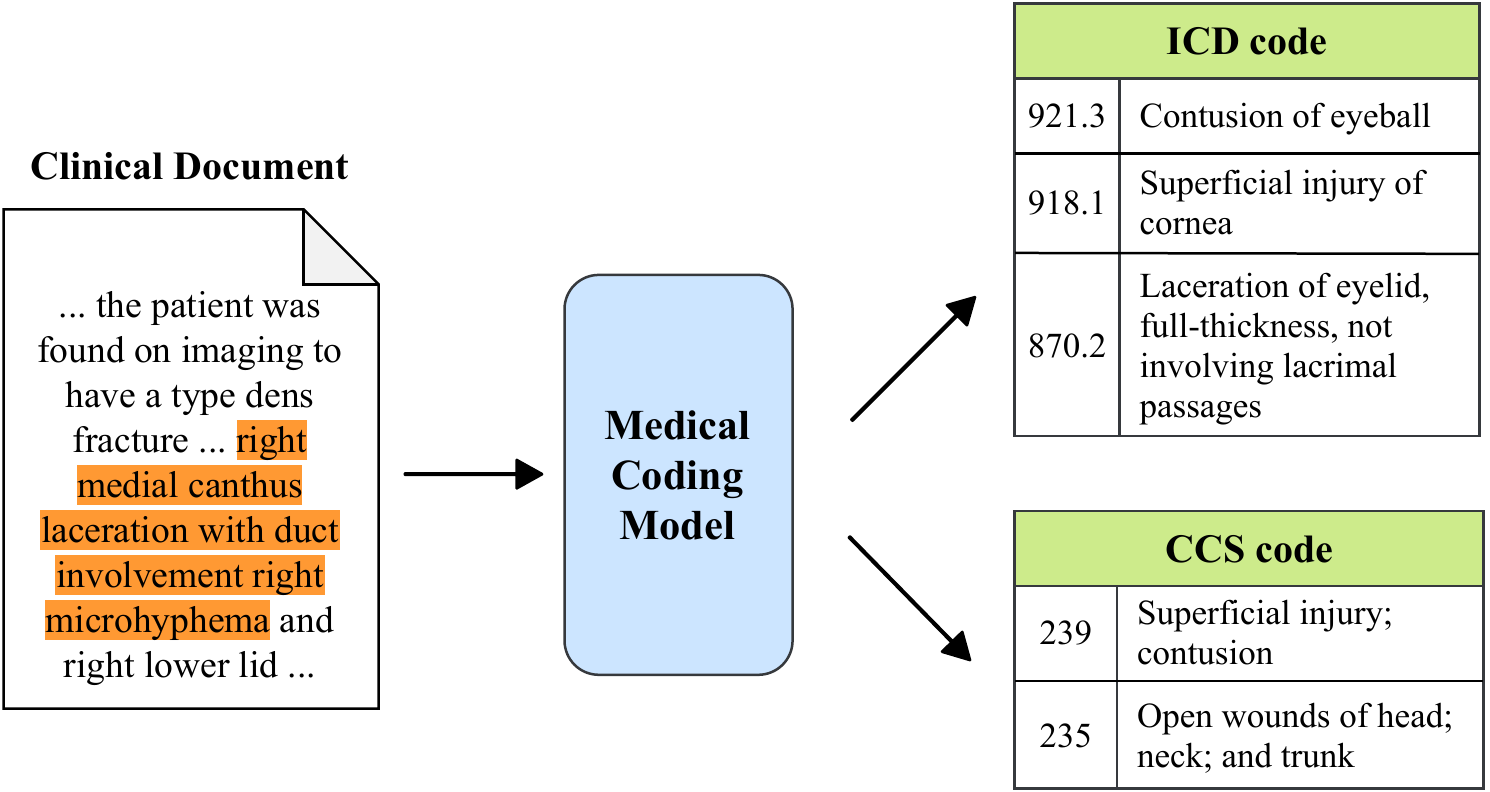}
	\caption{An example of medical code prediction, where ICD and CCS codes are used as the coding systems. The second column of each tables shows the disease name corresponding to each medical code.}
	\label{fig:example}
\end{figure*}

This paper primarily studies ICD and CCS coding systems because of their individual characteristics of popularization and simplicity.
CCS codes maintained by the Healthcare Cost and Utilization Project (HCUP\footnote{\url{www.hcup-us.ahrq.gov/toolssoftware/ccs/ccs.jsp}}) provide medical workers, insurance companies, and researchers with an easy-to-understand coding scheme of diagnoses and processes.
On the other hand, the ICD coding system provides a comprehensive classification tool for diseases and related health problems.
Nonetheless, the CCS and ICD codes have a one-to-many relationship that enables the CCS software to convert ICD codes into CCS codes with a smaller label space at different levels.
For instance, in Fig.~\ref{fig:example}, the ICD-CCS mapping scheme converts ``921.3'' (``Contusion of eyeball'') and ``918.1'' (``Superficial injury of cornea'') to the same CCS code ``239" , which represents the ``Superficial injury; contusion". The CCS code ``239" establishes a connection between two different ICD codes. 

Medical codes concisely summarize useful information from vast amounts of inpatient discharge summaries, and have high medical and commercial value.
They are consequently of interest for both medical institutions and health insurance companies.
For example, major insurance companies use standard medical codes in their insurance claim business~\cite{bottle2008intelligent}.
Professional coders do the medical coding task by annotating clinical texts with corresponding medical codes. 
Since manual coding is error-prone and labor-consuming~\cite{o2005measuring}, automated coding is needed. 
Taking the ICD coding as an example, many publications have proposed automated coding approaches, including feature engineering-based machine learning methods~\cite{perotte2014diagnosis,koopman2015automatic} and deep learning methods~\cite{cao2020hypercore,mullenbach2018explainable,li2020icd}.

However, the automated medical coding task is still challenging as reflected in the following two aspects. 
Clinical notes contain noisy information, such as spelling errors, irrelevant information, and incorrect wording, which may have an adverse impact on representation learning,  increasing the difficulty of medical coding. 
Also, it is a challenge to benefit from the relationship between different medical codes, especially when the label is high-dimensional. 
Existing automatic ICD coding models, such as CAML~\cite{mullenbach2018explainable} and MultiResCNN~\cite{li2020icd}, have limited performance because they do not consider the relationship between ICD codes. 
In the medical ontology, there exists certain connections between different concepts.
For example, in the ICD coding system, ``921.3" and ``918.1", representing ``Contusion of eyeball" and ``Superficial injury of cornea", respectively, belong to ``Superficial injury; contusion".
Medical coding models may suffer from underperformance if they can not effectively capture the relationships between medical codes.
For example in Fig.~\ref{fig:example}, the highlight area in a clinical document is converted into corresponding medical codes, including ICD codes and CCS codes.

In this paper, we propose a novel framework called MT-RAM, which combines \textbf{M}ulti\textbf{T}ask (MT) learning with a \textbf{R}ecalibrated \textbf{A}ggregation \textbf{M}odule (RAM) for medical code prediction. 
In particular, the RAM improves the quality of representation learning of clinical documents, by injecting rich contextual information and performing nested convolutions, thereby solving the challenge of encoding noisy and lengthy clinical notes.
In multitask training, we consider the joint training on two tasks, ICD and CCS code prediction. 
\textbf{M}ulti\textbf{T}ask \textbf{L}earning (MTL) is inspired by human learning, where people often apply the knowledge from previous tasks to help with a new task~\cite{zhang2017survey}. 
It makes full use of the information contained in each task, shares information between related tasks through common parameters, and enhances training efficiency~\cite{chandra2016evolutionary,yosinski2014transferable}. 
In addition, MTL reduces over-fitting to specific tasks by regularizing the learned representation to be generalizable across tasks~\cite{liu2019multi}. 
In the context of the two medical coding systems, CCS coding can promote the training on the ICD codes; further, the CCS codes can inform about the relationship between the ICD codes, thereby improving model performance. 

Our contributions fall into the following four aspects. 
\begin{itemize}
    \item To the best of our knowledge, this paper is the first to adopt multitask learning for medical code prediction and demonstrate the benefits of leveraging multiple coding schemes.
    \item We design a recalibrated aggregation module (RAM) to generate clinical document features with better quality and less noise.
    \item We propose a novel framework called MT-RAM, which combines multitask learning, bidirectional GRU, RAM and label-aware attention mechanism.
    \item Experimental results show competitive performance of our framework across different evaluation criteria on the standard real-world MIMIC-III database when compared with several strong baselines.
\end{itemize}

Our paper is organized as follows: Section~\ref{sec:related} introduces related work; Section~\ref{sec:method} describes the proposed model; Section~\ref{sec:exp} performs a series of comparison experiments, an ablation study and a detailed analysis of the properties of the RAM; finally, Section~\ref{sec:conclusion} provides concluding remarks.

\section{Related Work}
\label{sec:related}
\paragraph{Automated Medical Coding}
Automated medical coding is an essential and challenging task in medical information systems~\cite{perotte2014diagnosis}. 
Healthcare institutes use different medical coding systems such as ICD, one of the most widely used coding schemes.
The majority of early automated medical coding works use machine learning algorithms. 
Larkey and Croft~\cite{larkey1996combining} proposed a ICD code classifier with multiple models, including K-nearest neighbor, relevance feedback, and Bayesian independence classifiers. 
Perotte et al.~\cite{perotte2014diagnosis} presented two ICD coding approaches: a flat and a hierarchy-based SVM classifier. 
The experiments showed that hierarchical SVM model outperforms flat SVM because it captures the hierarchical structure of ICD codes.

Neural networks have gained popularity for medical coding with the recent advances of deep learning techniques. 
Recurrent neural networks capture the sequential nature of medical text and have been applied by several studies such as the attention LSTM~\cite{shi2017towards}, the Hierarchical Attention Gated Recurrent Unit (HA-GRU)~\cite{baumel2017multi}, and the multilayer attention-based bidirectional RNN~\cite{yu2019automatic}.
Convolutional networks also play an important role in this field.
Mullenbach et al.~\cite{mullenbach2018explainable} proposed Convolutional Attention network for Multi-Label classification (CAML). 
Li and Yu~\cite{li2020icd} utilized a Multi-Filter Residual Convolutional Neural Network (MultiResCNN), and Ji et al.~\cite{ji2020dilated} developed a dilated convolutional network.
Fine-tuning retrained language models as an emerging trend for NLP applications has been reported to have limits in medical coding by several initial studies~\cite{li2020icd,ji2020dilated} and a comprehensive analysis on the pretraining domain and fine-tuning architectures~\cite{ji2021does}.

\begin{figure*}[htbp]
\centering
	\includegraphics[width=\textwidth]{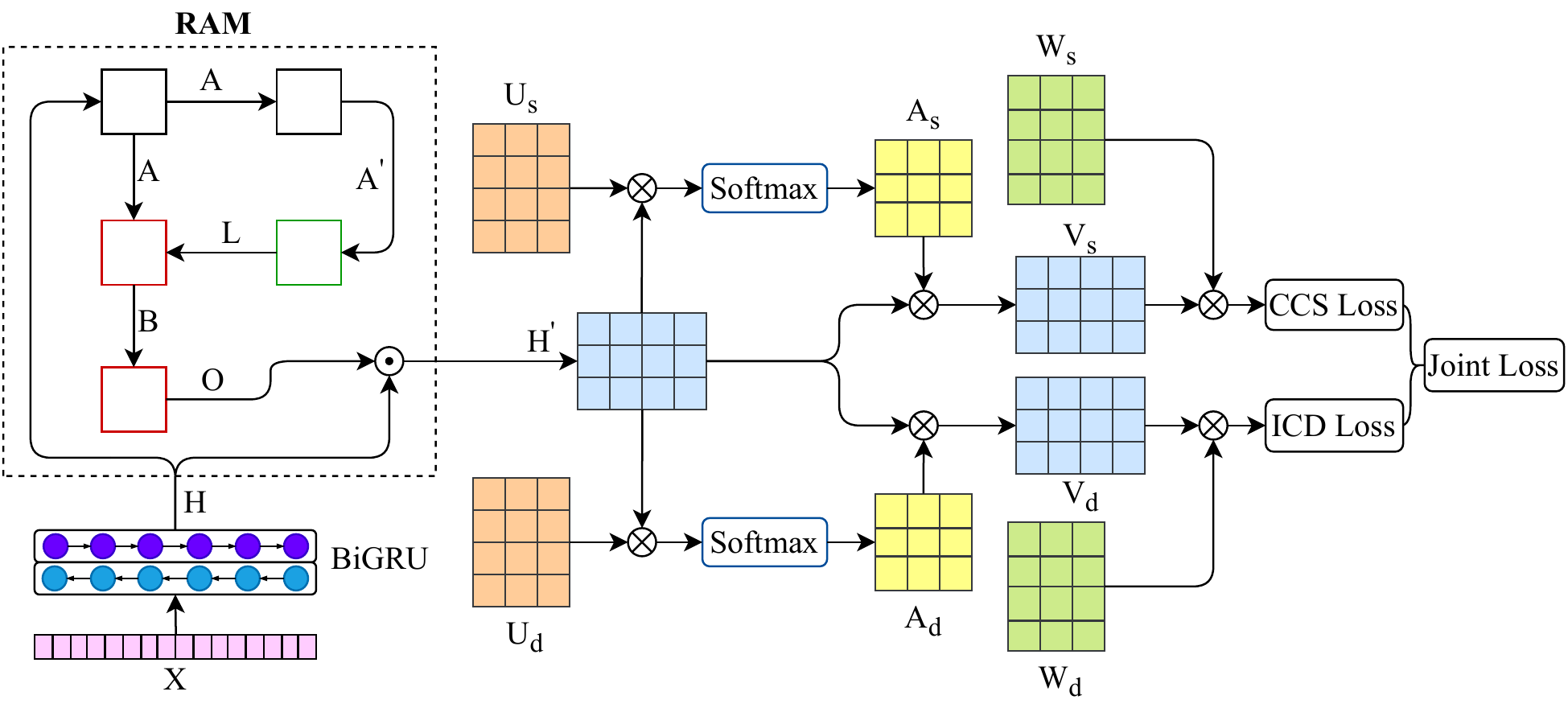}
	\caption{Model architecture. In the Recalibrated Aggregation Module (RAM), ``$\odot$" denotes as element-wise multiplication, ``\protect$\color{black} \medsquare$" represent a \textit{down node};
	`\protect$\color{green} \medsquare$" indicates a \textit{lateral node};
	``\protect$\color{red} \medsquare$" represents a \textit{up node}; 
	``$\otimes$" is the matrix multiplication operation.}

	\label{fig:model_overview}
\end{figure*}

\paragraph{Multitask Learning} 
Multitask learning is a machine learning paradigm that jointly trains multiple related tasks to improve the performance of each task and the generalization of the model. 
Multitask learning is widely used in various medical applications such as drug action extraction~\cite{zhou2018position}, biological image analysis~\cite{zhang2016deep} and clinical information extraction~\cite{bi2008improved,suk2016deep}. 
In recent years, researchers have studied leveraging multitask learning strategies to better process medical notes. 
Malakouti et al.~\cite{malakouti2019hierarchical} jointly trained different diagnostic models to improve performance of each diagnostic task. 
This work implemented the parameter sharing between tasks by utilizing the bottom-up and top-down steps.
This multitask learning framework improved the performance and the generalization ability of independently learned models.
Si and Roberts~\cite{si2019deep} presented a CNN-based multitask learning network for inpatient mortality prediction task, which comprises some related tasks such as 0-day, 30-day, 1-year patient death prediction.

\section{Method}
\label{sec:method}
This section describes the proposed {M}ulti-{T}ask {R}ecalibrated {A}ggregation {N}etwork, referred as \textbf{MT-RAM}, as it combines the \textbf{M}ulti-\textbf{T}ask learning scheme and a \textbf{R}ecalibrated \textbf{A}ggregation \textbf{M}odule.
The overall architecture of our MT-RAM network has five parts as shown in Fig.~\ref{fig:model_overview}.
We use word embeddings pretrained by the word2vec~\cite{mikolov2013distributed} as the input. 
Secondly, we use the bidirectional gated recurrent unit (BiGRU)~\cite{cho2014learning} layer to extract document representation features capturing sequential dependencies in clinical notes.
Next, a RAM module is used to improve the quality of the feature matrix and the efficiency of training for the multitask objective.
Fourthly, the attention classification layers with two branches of ICD and CCS codes are composed of label-wise attention mechanism and linear classification layers. 
The last part combines the respective losses of the two classification heads and performs multitask training.

\subsection{Input Layer}

Denote a clinical document with $n$ tokens as $w$ = \{$w_1, w_2, \dots, w_n$\}. We utilize word2vec \cite{mikolov2013distributed} to pretrain each clinical document to obtain word embedding matrices. 
A word embedding matrix, referred to $\mathbf{X} = [\mathbf{x}_1, \mathbf{x}_2, \dots, \mathbf{x}_n]^{\operatorname{T}}$, is the combination of each word vector $\mathbf{x}_n \in \mathbb{R}^{d_e}$, where $d_e$ is the embedding dimension. 
Next, we feed word embedding matrix $\mathbf{X} \in \mathbb{R}^{n \times d_e}$ into the BiGRU layer to extract document representation features.

\subsection{Bidirectional GRU Layer}
\label{sec:BiGRU}

We use a bidirectional GRU layer to extract the contextual information from the word embeddings $\mathbf{X}$ of the input documents. 
We calculate the latent states of GRUs on $i$-th token $x_i$:
\begin{align}
\overrightarrow{\mathbf{h}_i} &= \overrightarrow{\operatorname{GRU}}(\mathbf{x}_i, \overrightarrow{\mathbf{h}_{i-1}})\\
\overleftarrow{\mathbf{h}_i} &= \overleftarrow{\operatorname{GRU}}(\mathbf{x}_i, \overleftarrow{\mathbf{h}_{i+1}})
\end{align}
where $\overrightarrow{\operatorname{GRU}}$ and $\overleftarrow{\operatorname{GRU}}$ represent forward and backward GRUs, respectively. 
Final operation is to concatenate the $\overrightarrow{\mathbf{h}_i}$ and then $\overleftarrow{\mathbf{h}_i}$ into hidden vector $\mathbf{h}_i$:
\begin{align}
\mathbf{h}_i = \operatorname{Concat}(\overrightarrow{\mathbf{h}_i}, \overleftarrow{\mathbf{h}_i})
\end{align}
Dimension of forward or backward GRU is set to $d_r$. 
Bidirectional hidden vectors $\mathbf{h}_i\in \mathbb{R}^{2d_r}$ are horizontally concatenated into a resulting hidden representation  matrix $\mathbf{H} = [\mathbf{h}_1, \mathbf{h}_2, \dots, \mathbf{h}_n]^T$, where the dimension of $\mathbf{H} \in \mathbb{R}^{n \times 2d_r}$. 

\subsection{Recalibrated Aggregation Module}
\begin{figure}[htbp]
\centering
\includegraphics[width=\linewidth]{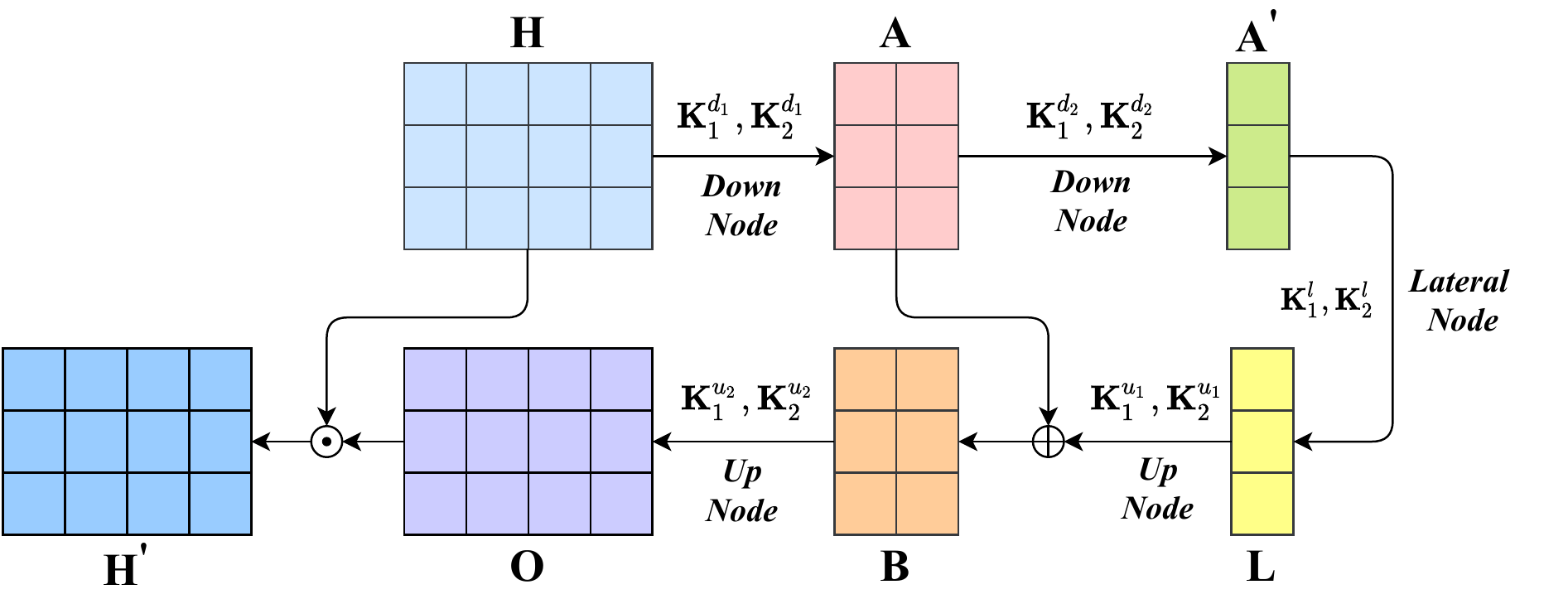}
\caption{The calculation flow of the RAM}
\label{fig:Matrix_flow}
\end{figure}

We propose a \textbf{R}ecalibrated \textbf{A}ggregation \textbf{M}odule (RAM) that abstracts features learned by the BiGRU, recalibrates the abstraction, aggregates the abstraction and the recalibrated features, and eventually combines the new representation with the original one.
This way, the RAM module can reduce the effect of noise in the clinical notes and lead to improved representations for medical code classification. 
In detail, the RAM leverages a nested convolution structure to extract and aggregate contextual information, which is used to recalibrate the noisy input features.
In addition to this, through the convolutions, the RAM attains global receptive fields during feature extraction, which is complementary to the GRU-based recurrent structure described in Sec.~\ref{sec:BiGRU}.
With these two characteristics, our RAM can improve the encoding of noisy and lengthy clinical notes.
The RAM consists of feature aggregation and recalibration.
The calculation flow of RAM is shown in Fig.\ref{fig:Matrix_flow} and described as below.

Firstly, the hidden representation $\mathbf{H}$ from the BiGRU layer passes through two down nodes to obtain matrices  $\mathbf{A}$ and $\mathbf{A^\prime}$. 
This downsampling process can be denoted as:
\begin{equation}\label{eq:downsample}
\mathbf{A} =  \bigwedge_{n=1}^{d_r} \Bigg\{ \mathbf{K}_2^{d_1} \left[ \operatorname{tanh}\left( \bigwedge_{m=1}^{d_r}  \left( \mathbf{K}_1^{d_1} \mathbf{H} \right)_m \right) \right]_n \Bigg\} \in \mathbb{R}^{n \times d_r},
\end{equation}
where $\bigwedge_{m=1}^{d_r}$ represents dislocation addition, i.e., the second matrix is shifted by one unit to the right based on the position of the first matrix. 
The overlapped area is summed up. 
We repeat this operation until the last matrix and cut off unit vectors on both sides of the concatenated matrix.
In Eq.~\ref{eq:downsample}, $\mathbf{K}_1^{d_1} \in \mathbb{R}^{2d_r \times k \times d_r}$  and $\mathbf{K}_2^{d_1} \in \mathbb{R}^{d_r \times k \times d_r}$ represent two convolutional kernel groups in the first down node and $k$ is the kernel size. 
The second downsampled matrix $\mathbf{A^\prime} \in \mathbb{R}^{n \times \frac{d_r}{2}}$ can also be obtained in a similar way with different convolutional kernel groups $\mathbf{K}_1^{d_2} \in \mathbb{R}^{d_r \times k \times \frac{d_r}{2}}$ and $\mathbf{K}_2^{d_2} \in \mathbb{R}^{\frac{d_r}{2} \times k \times \frac{d_r}{2}}$. 
Next, we use a lateral node with another two convolutional kernel groups $\mathbf{K}_1^{l} \in \mathbb{R}^{\frac{d_r}{2} \times k \times \frac{d_r}{2}}$ and $\mathbf{K}_2^{l} \in \mathbb{R}^{\frac{d_r}{2} \times k \times \frac{d_r}{2}}$, which have consistent in and out channel dimensions, to transform $\mathbf{A^\prime}$ into lateral feature matrix $\mathbf{L} \in \mathbb{R}^{n \times \frac{d_r}{2}}$.
We recover $\mathbf{L}$ with a up node and pair-wisely add the recovered signal with the first downsampled feature matrix $\mathbf{A}$ to obtain the primarily aggregated matrix $\mathbf{B}  \in \mathbb{R}^{n \times d_r}$ as denoted in Eq.~\ref{eq:addition}, where $\mathbf{K}_1^{u_1} \in \mathbb{R}^{\frac{d_r}{2} \times k \times d_r}$ and $\mathbf{K}_2^{u_1} \in \mathbb{R}^{d_r \times k \times d_r}$ represent deconvolutional kernel groups in the first up node. 
\begin{equation}\label{eq:addition}
\mathbf{B} = \mathbf{A} + \bigwedge_{n=1}^{d_r} \Bigg\{ \mathbf{K}_2^{u_1} \left[ \operatorname{tanh}\left( \bigwedge_{m=1}^{d_r} \left( \mathbf{K}_1^{u_1} \mathbf{L} \right)_m \right) \right]_n \Bigg\}
\end{equation}

\begin{figure}[htbp]
\centering
\includegraphics[width=\linewidth]{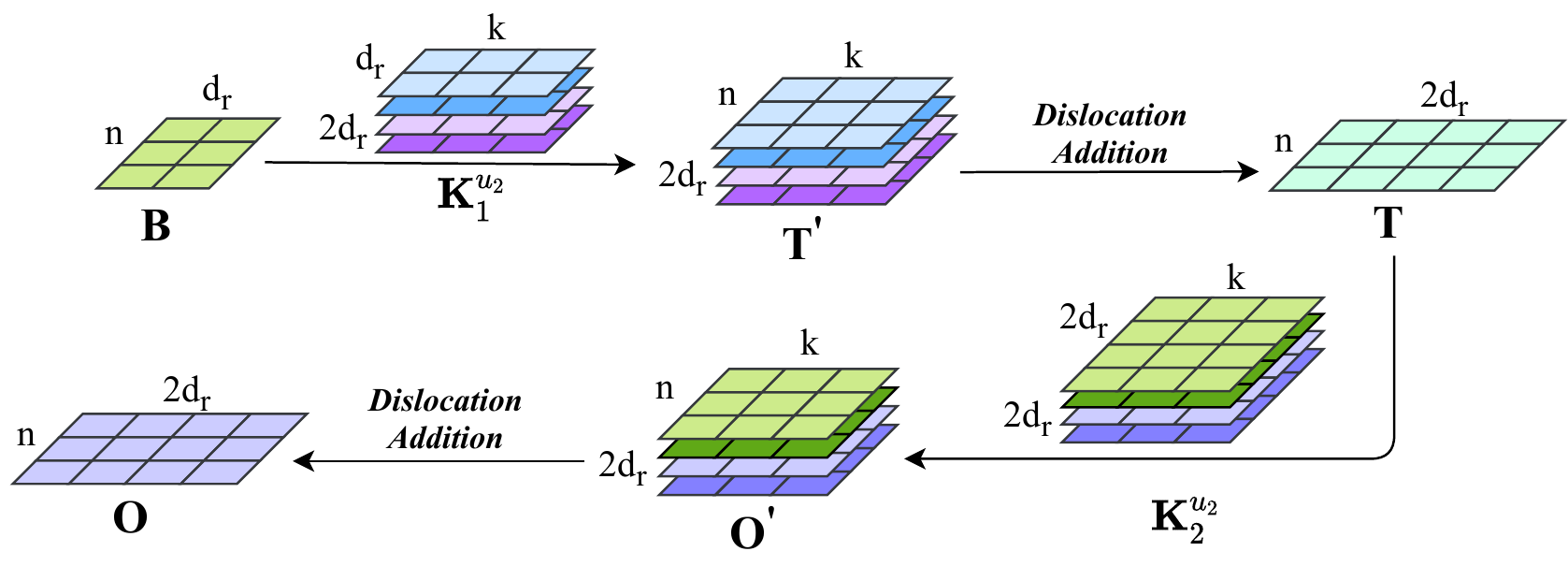}
\caption{Recover weight matrix $\mathbf{O}$ from the primary aggregation $\mathbf{B}$ in the RAM module. $k$ is the kernel size, and $d_r$ and $2d_r$ refer to the input and output feature dimensions in a up node.}
\label{fig:deconv}
\end{figure}

Secondly, we perform upsampling operations on the aggregated feature matrix $\mathbf{B}$ to obtain weight matrix $\mathbf{O} \in \mathbb{R}^{n \times 2d_r}$ as illustrated in Fig.~\ref{fig:deconv}.
Specifically, we leverage a deconvolution kernel group $\mathbf{K}_1^{u_2} \in \mathbb{R}^{d_r \times k \times 2d_r}$ 
to obtain the intermediate representation $\mathbf{T} \in \mathbb{R}^{n \times 2d_r}$. 
The different colors of $\mathbf{K}_1^{u_2}$ in Fig.~\ref{eq:deconv} correspond to how the different parts of the matrix $\mathbf{T^\prime} \in \mathbb{R}^{n \times k \times 2d_r}$ are calculated. This process is denoted as:
\begin{equation}
\label{eq:intermediate}
\mathbf{T} =  \bigwedge_{m=1}^{2d_r} \mathbf{T}_m^\prime = \bigwedge_{m=1}^{2d_r} (\mathbf{B} \mathbf{K}_1^{u_2})_m.
\end{equation}
We adopt a deconvolution operation on the intermediate representation $\mathbf{T}$ to get the weight matrix $\mathbf{O} \in \mathbb{R}^{n \times 2d_r}$, denoted as:
\begin{equation}\label{eq:deconv}
\mathbf{O} = \bigwedge_{n=1}^{2d_r} \mathbf{O}_n^\prime = \bigwedge_{n=1}^{2d_r} \left( \operatorname{tanh} \left(\mathbf{T}\right) \mathbf{K}_2^{u_2}\right)_n,
\end{equation}
where $\mathbf{K}_2^{u_2}$ represents the deconvolution kernel group.

Finally, we employ the feature recalibration in a way similar to the attention mechanism, where the ``attention'' score is learned by an iterative procedure with convolutional feature abstraction (Eq.~\ref{eq:downsample}) and de-convolutional feature excitation (Eq.~\ref{eq:deconv}).
Specifically, we multiply the input feature matrix $\mathbf{H}$ by the weight matrix $\mathbf{O}$ to obtain the recalibrated feature matrix $\mathbf{H^\prime} \in \mathbb{R}^{n \times 2d_r}$, denoted as:
\begin{equation}
\mathbf{H^\prime} = \operatorname{tanh} \left(\mathbf{O} \odot \mathbf{H}\right),
\end{equation}
where ``$\odot$" represents element-wise multiplication.
The recalibration operation enhances the original features with contextual information injection through the weight matrix $\mathbf{O}$, which comprises rich semantic information that is consequently less sensitive to errors. 
It enables the RAM module to have improved generalization ability and, in the end, improved performance in medical coding.

\subsection{Attention Classification Layers}
Features extracted by lower layers in shared modules are label-agnostic.
The Recalibrated Aggregation Module inherits the capacity of learning label-specific features from the Squeeze-and-Excitation block~\cite{hu2018squeeze} to some extent.
In order to make different positions of clinical notes correspond to different medical codes, we develop the label attention for classification layers to reorganize the characteristic information related to medical codes and enhance label specifications.
Working together with the RAM module and label attention mechanism, our model can achieve label-aware representation learning, which is helpful for multitask heads as described in the next section (Sec.~\ref{sec:multitask}).

The attention classification layers are described as follows. We take a subscript $d$ to denote a type of medical code. 
It can be generalized into different coding systems. 
Specifically, $d$ represents the ICD code in our paper. 
For simplicity, the bias term is omitted.
The attention scores of medical code $\mathbf{A}_d \in \mathbb{R}^{n \times m}$ can be calculated as:
\begin{equation}
\mathbf{A}_d = \operatorname{Softmax}(\mathbf{H^\prime}\mathbf{U}_d)
\end{equation}
where $\mathbf{H^\prime}$ is the document features extracted by the RAM block, $\mathbf{U}_d \in \mathbb{R}^{d_r \times m_d}$ represents the parameter matrix of query in the attention mechanism, and $m_d$ denotes the number of target medical code.
The attentive document features $\mathbf{V}_d \in \mathbb{R}^{d_r \times m_d}$ can be obtained by:
\begin{equation}
\mathbf{V}_d = \mathbf{A}_d^{\operatorname{T}} \mathbf{H^\prime}
\end{equation}
The label-wise attention mechanism captures the selective information contained in the document encoding $\mathbf{H^\prime}$ and the query matrix $\mathbf{U}_d$ determines what information in the encoding matrix to prioritize.

Then, we use a fully-connected max pooling layer as a classifier, 
which affines the weight matrix to obtain the score vector $\mathbf{Y}_d \in \mathbb{R}^{m_d \times 1}$ denoted as:
\begin{equation}
\mathbf{\mathbf{Y}}_d = \operatorname{Pooling}(\mathbf{W}_d\mathbf{V}_d^{\operatorname{T}})
\end{equation}
where $\mathbf{W}_d \in \mathbb{R}^{m_d \times m_d}$ represents the linear weight of the score vector.
We use the Sigmoid activation function to produce the probability logits $\mathbf{\bar{y}}_d$ for final prediction.

\subsection{Multitask Training}
\label{sec:multitask}

We introduce two self-contained tasks for multitask learning, i.e., ICD and CCS code prediction.
The two medical coding branch tasks enter different coding processes and back-propagate the ICD code loss and CCS code loss ,respectively. 
The structure of the two coding processing branches is similar.
By passing the encoded features of clinical notes through the label attention module, we can get the weighted document features of the ICD code $\mathbf{V}_d \in \mathbb{R}^{d_r \times m_d}$ and the CCS code $\mathbf{V}_s \in \mathbb{R}^{d_r \times m_s}$, where $m_d$ and $m_s$ is the number of ICD and CCS codes respectively. 
With the linear classifier layer, the prediction probability of ICD and CCS codes are generated as $\mathbf{\Bar{y}}_d$ and $\mathbf{\Bar{y}}_s $.

The medical code assignment is a typical multi-label classification task. 
We use the binary cross entropy loss as the loss function of each sub-task in the multitask setting. 
The ICD coding loss and CCS coding loss are denoted as:
\begin{align}
\mathcal{L}_d = \sum_{i=1}^{m_d} \Big[-y_{d_i}\log(\Bar{y}_{d_i}) - (1 - y_{d_i})\log(1 - \Bar{y}_{d_i}) \Big]\\
\mathcal{L}_s = \sum_{i=1}^{m_s} \Big[-y_{s_i}\log(\Bar{y}_{s_i}) - (1 - y_{s_i})\log(1 - \Bar{y}_{s_i}) \Big]
\end{align}
where $y_{d_i}, y_{s_i} \in \{0,1\}$ are the target medical code labels. $\Bar{y}_{d_i}$ and $\Bar{y}_{s_i}$ represent prediction probability of ICD and CCS codes, and the number of ICD and CCS codes are denoted as $m_d$ and $m_s$ respectively.
We adopt joint training for the two medical coding losses to facilitate multitask learning. 
The joint training loss is defined as
\begin{align}
\mathcal{L}_M = \lambda_d \mathcal{L}_d + \lambda_s \mathcal{L}_s,
\end{align}
where $\lambda_d$ and $\lambda_s$ are scaling factors of ICD and CCS codes.

\section{Experiments}
\label{sec:exp}
We perform a series of experiments to validate the effectiveness of our proposed model on public real-world datasets. 
Source code is available at \url{https://github.com/VRCMF/MT-RAM}. 

\subsection{Datasets}
\paragraph{MIMIC-III (ICD)} The third version
of Medical Information Mart for Intensive Care (MIMIC-III)\footnote{\url{https://mimic.physionet.org/gettingstarted/access/}} is a large, open-access dataset consists of clinical data associated with above 40,000 inpatients in critical care units of the Beth Israel Deaconess Medical Center between 2001 and 2012~\cite{johnson2016mimic}. 
Following Mullenbach et al.~\cite{mullenbach2018explainable} and Li and Yu~\cite{li2020icd}, we segment all discharge summaries documents based on the patient IDs, and generate 50 most frequent ICD codes for experiments. 
We refer MIMIC-III dataset with top 50 ICD codes as the MIMIC-III ICD dataset. 
There are 8,067 discharge summaries for training, and 1,574 and 1,730 documents for validation and testing, respectively. 
\paragraph{MIMIC-III (CCS)}
We utilize the ICD-CCS mapping scheme, provided by the HCUP, to convert the ICD codes and obtain the dataset with CCS codes. 
The converted CCS dataset denotes as MIMIC-III CCS, which contains 38 frequent CCS labels.
Because the MIMIC-III ICD dataset shares the discharge summary documents with the CCS dataset, the documents used for CCS code training, validation and testing are consistent with the ICD code documents. 
We change several conflicting mapping items so that ICD and CCS codes can achieve one-versus-one matching.
The converted CCS codes are then used as the labels of discharge summaries.

\subsection{Settings}
\paragraph{Data Preprocessing}
Following the processing flow of CAML~\cite{mullenbach2018explainable}, the non-alphabetic tokens, such as punctuation and numbers, are removed from clinical text. 
All tokens are transformed into lowercase format, and we replace low-frequency tokens (appearing in fewer than three documents) into the `UNK' token. 
We train the word2vec~\cite{mikolov2013distributed} on all discharge summaries to obtain the word embeddings. 
The maximum length of each document is limited to 2,500, i.e., documents longer than this length are truncated. The kernel size of convolution layer in the RAM module is 3.

\paragraph{Evaluation Metrics}

To evaluate the performance of models in CCS and ICD code (collectively called medical code) datasets, we follow the evaluation protocols of previous works~\cite{mullenbach2018explainable,li2020icd}. 
We utilize micro-averaged and macro-averaged F1, micro-averaged and macro-averaged AUC (area under the receiver operating characteristic curve), precision at $k$ as the evaluation methods. 
Precision at $k$ (`P@$k$' in shorthand) is the proportion of $k$ highest scored labels in the ground truth labels. 
When calculating of micro-averaged scores, each clinical text and medical codes are treated as separate predictions. 
During the computing of macro-averaged metrics, we calculate the scores for each medical code and take the average of them. 
We run our model ten times and report the mean and standard deviation of all the metrics. 

\paragraph{Hyper-parameter Tuning}
We refer to the previous works~\cite{mullenbach2018explainable,li2020icd} and apply some common hyper-parameter settings. 
Specifically, we set the word embedding dimension to 100, the maximum document length to 2500, dropout rate to 0.2, the batch size to 16, and the dimension of hidden units to 300.
In the choice of learning rate, 0.008 is the optimal learning rate, which achieves good model performance and consumes moderate time to converge. 
We set the scaling factors $\lambda_d$ and $\lambda_s$ to 0.7 and 0.3 respectively. 
We use different optimizers to train our model, including Adam~\cite{kingma2014adam}, AdamW~\cite{loshchilov2017decoupled} and SGD+momemtum~\cite{sutskever2013importance}. 
Although the AdamW optimizer can shorten the training time, its predictive performance is not as good as the Adam. 
The performance of the SGD+momentum and the Adam are close, while Adam converges faster.

\subsection{Baselines}

\noindent \textbf{CAML}~\cite{mullenbach2018explainable} comprises a single convolutional backbone and a label-wise attention mechanism, achieving high performance for ICD code prediction.

\noindent \textbf{DR-CAML}~\cite{mullenbach2018explainable}, i.e., the Description Regularized CAML, is an extension of CAML that incorporates the ICD description to regularize the CAML model.

\noindent \textbf{HyperCore}~\cite{cao2020hypercore}
uses the hyperbolic representation space to leverage the code hierarchy and utilize the graph convolutional network to capture the ICD code co-occurrence correlation.

\noindent \textbf{MultiResCNN}~\cite{li2020icd}
adopts a multi-filter convolutional layer to capture various text patterns and a residual connection to enlarge the receptive field.

\subsection{Results}
\label{sec:results}

\paragraph{\textbf{MIMIC-III (ICD codes)}}
Table~\ref{table:ICD} shows that the results of our MT-RAM model performs better than all baseline models on all evaluation metrics. When compared with the state-of-the-art MultiResCNN~\cite{li2020icd}, our model has improved the scores of macro-AUC, micro-AUC, macro-F1, micro-F1 and P@5 by 2.2\%, 1.5\%  4.5\%, 3.6\% and 2.3\% respectively. 
Our model outperforms the CAML~\cite{mullenbach2018explainable}, which is the classical automated ICD coding model, by 4.6\%, 3.4\%, 11.9\%, 9.2\% and 5.5\%. 
The improvement of our model in macro-F1 and micro-F1 is more significant than other metrics by comparing with HyperCore~\cite{cao2020hypercore}, specifically by 4.2\% and 4.3\% respectively. 
While other scores see moderate improvement by 1\% $\sim$ 3\%. 
Recent pretrained language models such as BERT~\cite{devlin2019bert} and its domain-specific variants like ClinicalBERT~\cite{alsentzer2019publicly} are omitted from the comparison because these models are limited to process text with 512 tokens and have been reported with poor performance by two recent studies~\cite{li2020icd,ji2021does}.

\begin{table}[htbp]
\setlength\tabcolsep{4pt} %
\begin{center}
\begin{tabular*}{0.9\textwidth}{@{\extracolsep{\fill}} lrr|rr|r }
\toprule
\multicolumn{1}{l}{\multirow{2}{*}{Models}} & \multicolumn{2}{c}{AUC-ROC} & \multicolumn{2}{c}{F1} & \multirow{2}{3em}{P@5}   \\
& \multicolumn{1}{c}{Macro}       & \multicolumn{1}{c|}{Micro}      & \multicolumn{1}{c}{Macro}      & \multicolumn{1}{c|}{Micro}      &      \\ 
\midrule
\multicolumn{1}{l}{CNN} & 87.6 & 90.7 & 57.6 & 62.5 & 62.0\\
\multicolumn{1}{l}{CAML} & 87.5& 90.9& 53.2& 61.4& 60.9\\ 
\multicolumn{1}{l}{DR-CAML}& 88.4  & 91.6 & 57.6 & 63.3& 61.8\\ 
\multicolumn{1}{l}{HyperCore}& 89.5$\pm$0.3 &92.9$\pm$0.2& 60.9$\pm$0.1 & 66.3\textbf{$\pm$0.1} &63.2$\pm$0.2\\
\multicolumn{1}{l}{MultiResCNN}& 89.9$\pm$0.4 & 92.8$\pm$0.2& 60.6$\pm$1.1& 67.0$\pm$0.3& 64.1\textbf{$\pm$0.1}\\
\hline
\multicolumn{1}{l}{MT-RAM (ours)} & \textbf{92.1$\pm$0.1} & \textbf{94.3$\pm$0.1}  & \textbf{65.2$\pm$0.3} & \textbf{70.7}$\pm$0.2 & \textbf{66.4}$\pm$0.2 \\
\bottomrule
\end{tabular*}
\end{center}
\captionsetup{justification=centering}
\caption{MIMIC-III results (ICD code). Results are shown in \%. We set different random seeds for initialization to run our model for 10 times. Results of MT-RAM are demonstrated in \textit{means} $\pm$ \textit{standard deviation}}
\label{table:ICD}
\end{table}

\paragraph{\textbf{MIMIC-III (CCS code)}}
We evaluate the CAML, DR-CAML and the MultiResCNN on the MIMIC-III CCS dataset and record the results in Table~\ref{table:CCS}.
Since the Hypercore~\cite{cao2020hypercore} does not provide the source code, we omit it from the comparison.
Following the practice described in the section of hyper-parameter tuning, we set all the parameters of the CAML and the MultiResCNN to be consistent with the hyper-parameters of the original works except for the learning rate. 

As shown in Table~\ref{table:CCS}, we can see that our model obtains better results in the macro AUC, micro AUC, macro F1, micro F1, P@5, compared with the strong MultiResCNN baseline. 
The improvement of our model is 1.6\% in both macro AUC and micro AUC, 4.2\% in macro F1, 3.4\% in micro F1, and 2.7\% in P@5. 
DR-CAML uses the ICD code description to achieve performance improvement. DR-CAML uses the description of ICD codes to improve the performance of CAML. But on MIMIC-III (CCS) dataset, this description will cause interference to CAML, so the result of DR-CAML is worse.
Our model improves the F1 macro metric by 5.5\%, comparing with the CAML model.

\begin{table}[htbp]
\setlength\tabcolsep{4pt} %
\begin{center}
\begin{tabular*}{0.9\textwidth}{@{\extracolsep{\fill}} lrr|rr|r }
\toprule
\multicolumn{1}{l}{\multirow{2}{*}{Models}} & \multicolumn{2}{c}{AUC-ROC} & \multicolumn{2}{c}{F1} & \multirow{2}{3em}{P@5}   \\
& \multicolumn{1}{c}{Macro}       & \multicolumn{1}{c|}{Micro}      & \multicolumn{1}{c}{Macro}      & \multicolumn{1}{c|}{Micro}      &      \\ 
\midrule
\multicolumn{1}{l}{CAML}        & 89.2$\pm$0.3 	    & 92.2$\pm$0.3     & 60.9$\pm$0.9   & 67.5$\pm$0.4   &	64.5$\pm$0.4   \\ 

\multicolumn{1}{l}{DR-CAML}     & 87.5$\pm$0.4  &	90.5$\pm$0.4  &	59.3$\pm$1.0 &	65.6$\pm$0.6 &	62.6$\pm$0.5 \\ 

\multicolumn{1}{l}{MultiResCNN}            & 89.2$\pm$0.2 &	92.4$\pm$0.2 &	62.9$\pm$0.9 &	68.8$\pm$0.6 &	64.6$\pm$0.3  \\
\hline
\multicolumn{1}{l}{MT-RAM (ours)} & \textbf{92.2$\pm$0.1} & \textbf{94.6$\pm$0.1} & \textbf{69.4$\pm$0.1} & \textbf{74.4$\pm$0.2} & \textbf{68.4$\pm$0.1} \\
\bottomrule
\end{tabular*}
\end{center}
\captionsetup{justification=centering}
\caption{MIMIC-III results (CCS code). We run each model for 10 times and each time set different random seeds for initialization. Results of all models are demonstrated in \textit{means} $\pm$ \textit{standard deviation}}
\label{table:CCS}
\end{table}

\subsection{Ablation Study}
We examine the general usefulness of the two main components - multitask training (MTL) and RAM module, by conducting an ablation study, where we consider the performance of three representative ICD coding models: CAML, MultiResCNN, and the GRU-based model (our method), with and without the specific components.

\paragraph{\textbf{Multitask Learning}}
We firstly investigate the effectiveness of the multitask learning (MTL) scheme.
From Table~\ref{table:modelMM}, we can observe that CAML and BiGRU have been improved by a relatively large margin across all evaluation metrics with multitask training. 
The CAML with MTL achieves 7.6\% and 5.2\% improvement in macro and micro F1, respectively, and obtains increases by about 2\% to 3\% in other scores. 
Similarly, the BiGRU with MTL has achieved a good improvement in macro and micro F1, increased by 4.2\% and 3.3\% respectively. 
For the MultiResCNN model, the multitask learning also contributes to relatively good results, which is 2.3\% improvement in the macro F1 score. 
The reason why multitask learning can improve the performance of the model is the information exchange between the two tasks. 
Intuitively, there exists a correlation relationship between ICD and CCS coding systems
This leads to complementary benefits for both ICD and CCS code prediction tasks.
CAML and MultiResCNN have achieved significant gains by incorporating the multitask learning aggregation framework as a whole, i.e., the multitask learning scheme and the RAM together. 
Therefore, the gain of the multitask learning aggregation framework is not limited to some special network structures, and it has strong generalization ability.

\begin{table}[htbp]
\setlength\tabcolsep{4pt} %
\begin{center}
\begin{tabular*}{0.9\textwidth}{@{\extracolsep{\fill}} lrr|rr|r }
\toprule
\multicolumn{1}{l}{\multirow{2}{*}{Models}} & \multicolumn{2}{c}{AUC-ROC} & \multicolumn{2}{c}{F1} & \multirow{2}{3em}{P@5}   \\
& \multicolumn{1}{c}{Macro}       & \multicolumn{1}{c|}{Micro}      & \multicolumn{1}{c}{Macro}      & \multicolumn{1}{c|}{Micro}      &      \\ 
\midrule
\multicolumn{1}{l}{CAML}                   & 87.5                     & 90.9                      & 53.2                     & 61.4                      & 60.9                     \\
\multicolumn{1}{l}{CAML + RAM}               & 91.3                     & 93.5                      & 61.4                     & 67.4                      & 65.1                    \\
\multicolumn{1}{l}{CAML + MTL}               & 90.8                     & 93.2                      & 60.8                     & 66.6                      & 64.0                     \\
\multicolumn{1}{l}{CAML + MTL + RAM}           & \textbf{91.4}                     & \textbf{93.8}                      & \textbf{62.5}                     & \textbf{68.7}                      & \textbf{65.3}                     \\ 
\midrule
\multicolumn{1}{l}{MultiResCNN}            & 89.9                     & 92.8                      & 60.6                     & 67.0                      & 64.1                     \\
\multicolumn{1}{l}{MultiResCNN + RAM}               & 91.2                     & 93.4                     & 62.4                     & 68.1                      & 64.7                     \\
\multicolumn{1}{l}{MultiResCNN + MTL}        & 90.8                     & 93.2                      & 62.9                     & 67.8                      & 64.3                     \\
\multicolumn{1}{l}{MultiResCNN + MTL + RAM}    & \textbf{91.7}                     & \textbf{93.9}                      & \textbf{64.1}                     & \textbf{69.0}                      & \textbf{65.0}                     \\ 
\midrule
\multicolumn{1}{l}{BiGRU}               & 91.0 & 93.4& 60.4 & 66.6 &64.4 \\
\multicolumn{1}{l}{BiGRU + RAM}           & 91.7                     & 93.6                      & 63.5                     & 69.1                      & 65.0                     \\
\multicolumn{1}{l}{BiGRU + MTL}           & 91.8                     & 94.1                      & 64.6                     & 69.9                      & 66.2                     \\
\multicolumn{1}{l}{BiGRU + MTL + RAM (MT-RAM)}       & \textbf{92.1}                     & \textbf{94.3}                      & \textbf{65.1}                     & \textbf{70.6}                      & \textbf{66.4}                     \\
\bottomrule
\end{tabular*}
\end{center}
\captionsetup{justification=centering}
\caption{Ablation study}
\label{table:modelMM}
\end{table}

\paragraph{\textbf{Recalibrated Aggregation Module}}
The second part of ablation study if examines whether the proposed Recalibrated Aggregation Module (RAM) can learn useful features and consequently lead to better performance. 
In Table~\ref{table:modelMM}, the performance of the three models has been greatly improved after including the RAM module to the multitask BiGRU architecture. 
The micro F1 scores of CAML, MultiResCNN and MT-RAM have been improved by 2.1\%, 1.2\% and 0.8\%, respectively. 
The RAM module helps the GRU-based model achieve greater improvement than convolution-based models.

\subsection{A Detailed Analysis of the Properties of the RAM}
We conduct an exploratory study to investigate the effectiveness of element-wise multiplication in the final feature weighting stage. 
We denote models applying the multitask learning and multiplicative  to CAML and MultiResCNN as MT-CAML + RAM (Mult) and MT-MultiResCNN + RAM (Mult), respectively.
The RAM (Add) means to replace the multiplication operation in RAM with an addition operation.
From Table~\ref{table:analy-RAM}, we can observe that the model with RAM (Mult) outperforms models with RAM (Add) in most evaluation metrics. 
Although the results of MT-RAM (Add) in F1 macro, F1 micro and P@5 are slightly better than the results of MT-RAM (Mult), the gap is marginal. Considering the generalization ability and performance improvement of the two modules, the RAM with multiplication operation outperforms the RAM with addition operation.

\begin{table}[htbp]
\setlength\tabcolsep{4pt} %
\begin{center}
\begin{tabular*}{0.88\textwidth}{@{\extracolsep{\fill}} lrr|rr|r }
\toprule
\multicolumn{1}{l}{\multirow{2}{*}{Models}} & \multicolumn{2}{c}{AUC-ROC} & \multicolumn{2}{c}{F1} & \multirow{2}{3em}{P@5}   \\
& \multicolumn{1}{c}{Macro}       & \multicolumn{1}{c|}{Micro}      & \multicolumn{1}{c}{Macro}      & \multicolumn{1}{c|}{Micro}      &      \\ 
\midrule
\multicolumn{1}{l}{MT-CAML + RAM (Add)}               & 91.1&	93.5&	62.1&	68.1&	65.0                     \\
\multicolumn{1}{l}{MT-CAML + RAM (Mult)}           & \textbf{91.4}                     & \textbf{93.8}                      & \textbf{62.5}                     & \textbf{68.7}                      & \textbf{65.3}                     \\ 
\midrule
\multicolumn{1}{l}{MT-MultiResCNN + RAM (Add)}        & 91.1&	93.3&	62.4&	67.7&	64.1                    \\
\multicolumn{1}{l}{MT-MultiResCNN+ RAM (Mult)}    & \textbf{91.7}                     & \textbf{93.9}                      & \textbf{64.1}                     & \textbf{69.0}                      & \textbf{65.0}                     \\ 
\midrule
\multicolumn{1}{l}{MT-RAM (Add)}           & 92.0&	94.1&	\textbf{65.9}&	\textbf{70.8}&	\textbf{66.7} \\
\multicolumn{1}{l}{MT-RAM (Mult)}       & \textbf{92.1}                     & \textbf{94.3}                      & 65.1                     & 70.6                      & 66.4                     \\
\bottomrule
\end{tabular*}
\end{center}
\captionsetup{justification=centering}
\caption{Analysis of RAM: multiplicative versus additive}
\label{table:analy-RAM}
\end{table}

Regarding to the position of RAM in the multitasking learning framework, we found that it is best to embed the RAM in the shared layers.
Compared with putting RAM in the two branches of the framework, RAM module embedded in the shared layers helps two sub-tasks share more information.
If RAM is embedded in two sub-branch networks, the depth of the sub-network will increase and the shared part will decrease.
The deepening of the sub-network will interfere with the network convergence and make training more difficult.
At the same time, reducing the shared part will reduce the amount of information exchange between sub-tasks, which will affect the improvement of the model by the multitask learning scheme.

\section{Conclusion}
\label{sec:conclusion}
In this paper, we proposed a novel multitask framework for the automated medical coding task, which improved feature learning for clinical documents and accounted for the dependencies between different medical coding systems.
We designed a Recalibrated Aggregation Module (RAM) to enrich document features and reduce noisy information.
Furthermore, we leveraged multitask learning to share information across different medical codes.
We demonstrated that the combination of multitask learning and RAM improved automatic medical coding considerably. In addition, these components are generalizable and can be successfully integrated to other overall architectures.
The experimental results on the real-world clinical MIMIC-III database showed that our framework outperformed previous strong baselines.
Finally, we believe our framework can be beneficial not only in medical coding tasks, but also in other text label prediction tasks.

\section*{Acknowledgments}
This work was supported by the Academy of Finland (grant 336033) and EU H2020 (grant 101016775).
We acknowledge the computational resources provided by the Aalto Science-IT project.
The authors wish to acknowledge CSC - IT Center for Science, Finland, for computational resources.

%
%
\bibliographystyle{plain}
\bibliography{mt-coding.bib}

\end{document}